\documentclass{article}
\usepackage{spconf,amsmath,graphicx,enumitem}

\newcommand\RLis{%
  \raisebox{-.25ex}{\rotatebox[origin=c]{-90}{\includegraphics[height=10pt]{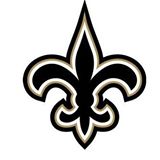}}}}
\setlist[itemize]{label=\RLis}
\usepackage{lscape}

\DeclareMathOperator*{\minimize}{minimize\quad}

\title{Son of Zorn's Lemma: Targeted Style Transfer Using Instance-aware Semantic Segmentation}
\name{Carlos Castillo, Soham De, Xintong Han, Bharat Singh, Abhay Kumar Yadav, and Tom Goldstein
\thanks{TG was supported in part by the US NSF under grant CCF-1535902 and by the US Office of Naval Research under grant N00014-17-1-2078. }
}
\address{Department of Computer Science, University of Maryland, College Park}
\begin{document}
\topmargin=0mm
%
\maketitle
\begin{abstract}
Style transfer is an important task in which the style of a source image is mapped onto that of a target image.  The method is useful for synthesizing derivative works of a particular artist or specific painting.  This work considers {\em targeted} style transfer,  in which the style of a template image is used to alter only {\em part} of a target image.  For example, an artist may wish to alter the style of only one particular object in a target image without altering the object's general morphology or surroundings.  This is useful, for example, in augmented reality applications (such as the recently released {\em Pok\'emon go}), where one wants to alter the appearance of a single real-world object in an image frame to make it appear as a cartoon. Most notably, the rendering of real-world objects into cartoon characters has been used in a number of films and television show, such as the upcoming series {\em Son of Zorn}.   
We present a method for targeted style transfer that simultaneously segments and stylizes single objects selected by the user.  The method uses a Markov random field model to smooth and anti-alias outlier pixels near object boundaries, so that stylized objects naturally blend into their surroundings. 


\end{abstract}
\section{Introduction}
\label{sec:intro}

\begin{figure}[t]
  \centering
    \includegraphics[width=.35\textwidth]{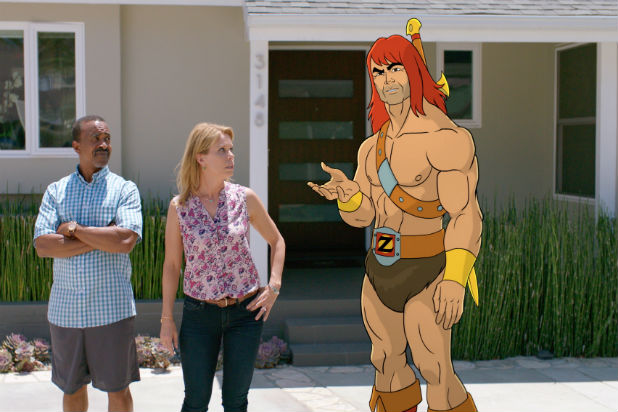}
    \vspace{-2mm}
  \caption{Son of Zorn is one recent example of a media production that uses cartoon textures implanted within a real-world setting. Image credited to FOX Broadcasting Company.}
  \label{fig:son-of-zorn}
\vspace{-3mm}
\end{figure}

\begin{figure*}[t]
  \centering
    \includegraphics[width=.9\textwidth]{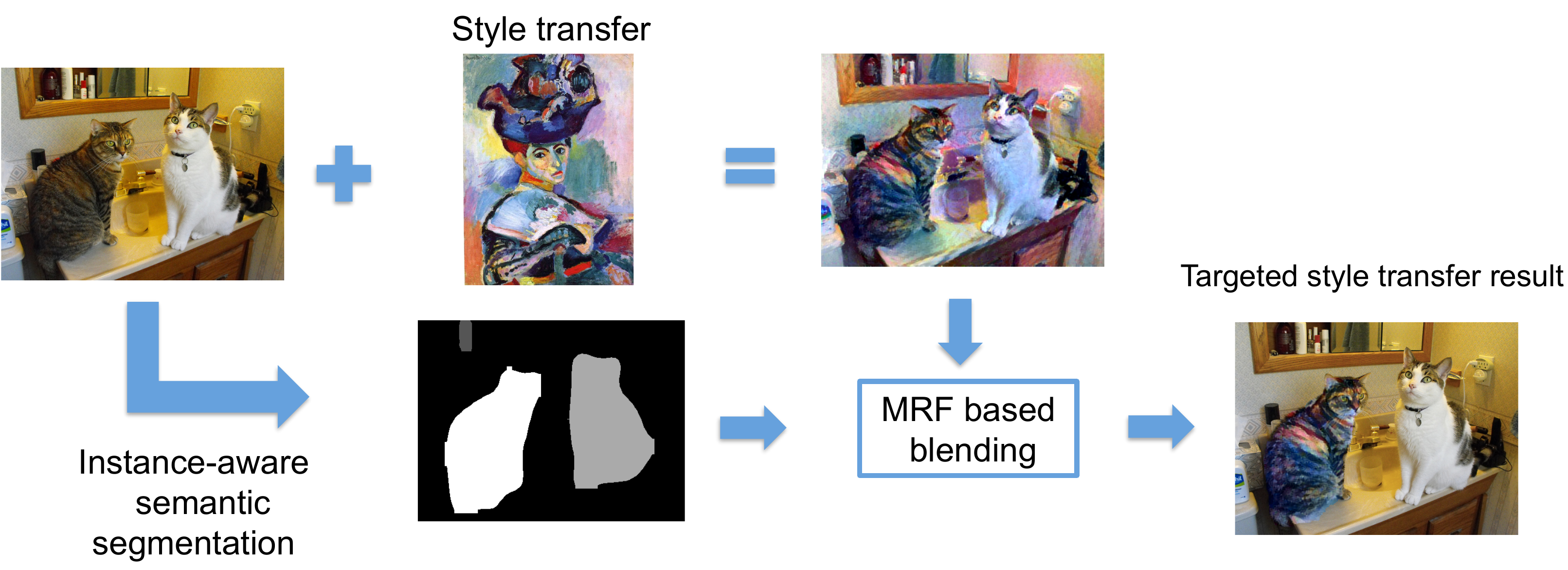}
    \vspace{-3mm}
  \caption{An overview of our algorithm.}
  \label{fig:overview}
\vspace{-3mm}
\end{figure*}

Style transfer is an important task in computer graphics in which the style (line stokes, textures, and colors) of a source image is mapped onto that of a target image.  Automated style transfer software facilitates the conversion of real-world images into the appropriate style to form the background in cartoons, simulations, and other renderings. The method is also useful for generating derivative works of a particular artist or painting.  The concept of style transfer is behind popular apps like {\em Prisma}, which convert real-world photos into different artistic styles.

In this paper, we propose {\em targeted} style transfer,  in which the style of a template image is used to alter only {\em part} of a target image.  For example, an artist may wish to alter the style of only one particular object in a target image without altering the object's general morphology or surroundings.  This is useful, for example, in augmented reality applications (such as the recently released {\em Pok\'emon go}), where one wants to alter the appearance of a single real-world object in an image frame to make it appear as a cartoon. Most notably, the rendering of real-world objects into cartoon characters has been used in a number of films and television show, such as the upcoming series {\em Son of Zorn} (see Fig. \ref{fig:son-of-zorn}).

We present a method for targeted style transfer that simultaneously segments and stylizes a single object selected by the user.  The method performs the object transformation using deep network-based image modification hybridized with semantic segmentation.  The method integrates a Markov random field model to smooth and anti-alias outlier pixels near object boundaries, so that stylized objects naturally blend into their surroundings without visible seams.  


\subsection{Related Work}
Whole-image style transfer has been studied by a number of authors.  Style transfer was first proposed in \cite{gatys2016image,gatys2015neural}, in which a deep convolutional network was used to transduce the style of a source image onto the target.  The transformed image is recovered by minimizing an energy functional with two terms.  The first term measures the semantic similarity between the target image and generated image, as quantified by the 2-norm difference between the deep features of each image.  The second term measures the texture similarity between the generated image and the source image.  Texture information is extracted from each image using covariance matrices that capture the correlations between deep features within an image.  This method of texture extraction was first proposed in \cite{gatys2015texture} for texture synthesis.

Several authors have proposed improvements to the style transfer model, although all have been focused on whole images. In \cite{gatys2016preserving}, the authors modify the transfer algorithm to be color preserving.  This is accomplished by modifying the source image to have a color profile similar to the target before performing the style mapping.  The authors of \cite{johnson2016perceptual} speed up style transfer by using a simplified ``perceptual loss function'' to compute the similarity between images.
%
In \cite{li2016combining},  the authors present a model for data-driven image synthesis that, given an image, automatically creates a variant that looks similar but differs in structure. 
%
The model uses a combination of generative Markov random fields and deep convolutional neural networks (dCNN) for synthesizing the images. 

Since this paper solves the problem of style transfer for a targeted object, our approach needs to generate a mask for each object in the target image. Therefore, it is also related to object detection \cite{girshick2015fast,ren2015faster,he2014spatial} and semantic segmentation \cite{pinheiro2015learning,dai2015instance}. Faster R-CNN \cite{ren2015faster} introduces a Region Proposal Network that predicts object boxes and objectness scores at the same time with an almost cost-free region proposal process. DeepMask \cite{pinheiro2015learning} trains a neural network with two objectives jointly. 
%
Also,  \cite{dai2015instance} proposes a cascaded network with three stages, which predict box instances, mask instances, and categorized instances in an end-to-end multi-tasking framework. In this paper, we utilize the method in \cite{dai2015instance} because it provides accurate mask instances for objects.

\section{Our Approach}
In this paper, we introduce a new pipeline for performing style transfer only on \emph{parts} of images. 
%
The pipeline of the basic algorithm is shown in Fig. \ref{fig:overview}:

\begin{itemize}
\item We first map the style of the source image onto the whole target image using the style transfer algorithm as described in \cite{gatys2016image}. 
\item 
A semantic segmentation algorithm \cite{dai2015instance} identifies different regions in the target image, and the user selects the regions onto which transfer will occur. The user may select specific objects in images, for example a specific person or a group of people in an image, to accept the style transfer.
\item The target object is segmented from the style transferred image, and a Markov random field (MRF) based model is used to merge the extracted stylized object with the non-stylized background. 
\end{itemize}

Note that a naive style transfer could be done by segmenting the stylized object and placing it into the non-stylized background without the MRF model.  The naive transfer yields a crude preliminary result, but the solution often looks out-of-place. The MRF model described below produces a more appealing embedding of the stylized object into the background. 
 To the best of our knowledge, this is the first paper to study a pipeline containing segmentation, style transfer, and image fusion.
We describe each step below.
\subsection{Style Transfer}
Our algorithm is built on the style transfer algorithm of Gatys et al. \cite{gatys2016image}.  Given a source image $s$ containing a prescribed style, and an target image $t,$ the algorithm recovers an image $x$ with deep features similar to the target image, but with texture information taken from $s.$  This is accomplished by solving the (highly) non-linear least-squares problem
$$\minimize_x \quad  \underbrace{\| F(x) - F(t)  \|^2}_{\text{structure}}  +  \underbrace{\| C(x) - C(s) \|^2}_{style}$$
where $F(\cdot)$ is a function mapping an image onto its deep features, and $C(\cdot)$ maps an image on a covariance matrix that measures the correlations of deep features in space.    For detailed construction of these operators, see \cite{gatys2016image}.  This problem is solved using back-propagation on $x$ as implemented in the popular deep learning library {\em Torch}. 

\subsection{Instance Segmentation}
 
We use an instance-aware semantic segmentation method \cite{dai2015instance} to generate a mask for each object instance in an image. Our interface enables a user to simply click on a semantic instance, and the image style is transferred to that instance.  The instance semantic segmentation approach is built on a cascaded multi-task network using the loss function:
\vspace{-2mm}
\begin{equation*} \small
L(\theta) = L_b(B(\theta)) + L_m(M(\theta) | B(\theta)) + L_c(C(\theta) | M(\theta), B(\theta)) 
\end{equation*}
where $\theta$ is the weight parameters of the neural network. There are three loss terms where the latter ones depend on the former ones. $L_b$ is the loss function of Region Proposal Networks (RPNs) introduced in \cite{ren2015faster}, which generates bounding box locations and predicts their ``objectness'' scores $B(\theta)$. $L_m$ is the loss of the second stage, where Region-of-Interest (RoI) pooling \cite{he2014spatial} is used to extract features in the predicted boxes and a binary logistic regression predicts the instance mask $M(\theta)$. Finally, as in \cite{hariharan2014simultaneous}, the softmax classification loss $L_c$ is computed on top of concatenated pathways of masks $M(\theta)$ and boxes $B(\theta)$, and the last stage outputs the class prediction scores $C(\theta)$ for all instances.


\subsection{Using MRFs to Blend Images}
To blend the targeted style transferred object (which call the foreground in this section) into the original image (background) smoothly, we use a Markov random field. We composite the stylized/foreground and original/background images by solving an optimization problem to choose among possible labels (either foreground or background) for each pixel in the image. The properties of an ideal blending are:
\begin{itemize}
  \item The boundary between stylized and original pixels should be near the original segmentation boundary. 
  \item The seams should not draw attention, i.e., the stylized object should blend smoothly into the background.
\end{itemize}

\sloppy
We formulate an objective function that approximately measures these properties,
and then minimize it using Markov Random Field (MRF) optimization to assign a foreground/background label to each pixel.  We first define a narrow band of ambiguous pixels near the foreground/background object boundary.  Points outside of this ambiguous band have their labels fixed to the value assigned during the original semantic segmentation.   Only points within this band will be adjusted to achieve a smooth effect.

For an ambiguous pixel with image coordinates $p
= (p_x, p_y),$ and label $l$ which can be either foreground or background, we
define the unary potentials $U(p,l)$ as:
\begin{align*}
  U(p,l) = \| p - c^l\|.
\end{align*}
where $c^l = (c^l_x, c^l_y)$ is the closest non-ambiguous pixel to $p$ in region $l$. This encourages the model to select a background label for pixels lying near the boundary between background and ambiguous pixels, and a foreground label for pixels lying near the foreground side of the boundary.

The binary potential term in the MRF encourages smooth transitions between the foreground and background.  Let $I_l(p)$ denote the intensity of the foreground image at pixel p when $l$ is the foreground label,  and the background intensity when $l$ is the background label.  Given two pixels, $p_1$ and $p_2$, with respective labels $l_1$ and $l_2$, we definite the pairwise energy
\begin{align*}
  B(p_1, l_1, p_2, l_2) = |I_{l_1}(p_1) - I_{l_2}(p_1)|^2 + |I_{l_2}(p_2) - I_{l_1}(p_2)|^2.
\end{align*}
This energy forces the transition between foreground/background to occur near pixels that are least effected by the stylization. 
Finally, we define the following energy function:
\begin{align*}
  E(l) = \sum_{p} U(p, l_p) + \sum_{\{p,q\} \in \mathcal{N}} B(p,l_p, q, l_q).
\end{align*}
where $\mathcal{N}$ contains sets of neighboring pixels. We obtain the labels $l$ by minimizing the energy $E(l)$ using gco-v3~\cite{boykov2001fast}.

\section{Experimental Results}
\begin{figure*}[]
   \centering
   \includegraphics[width=1\linewidth]{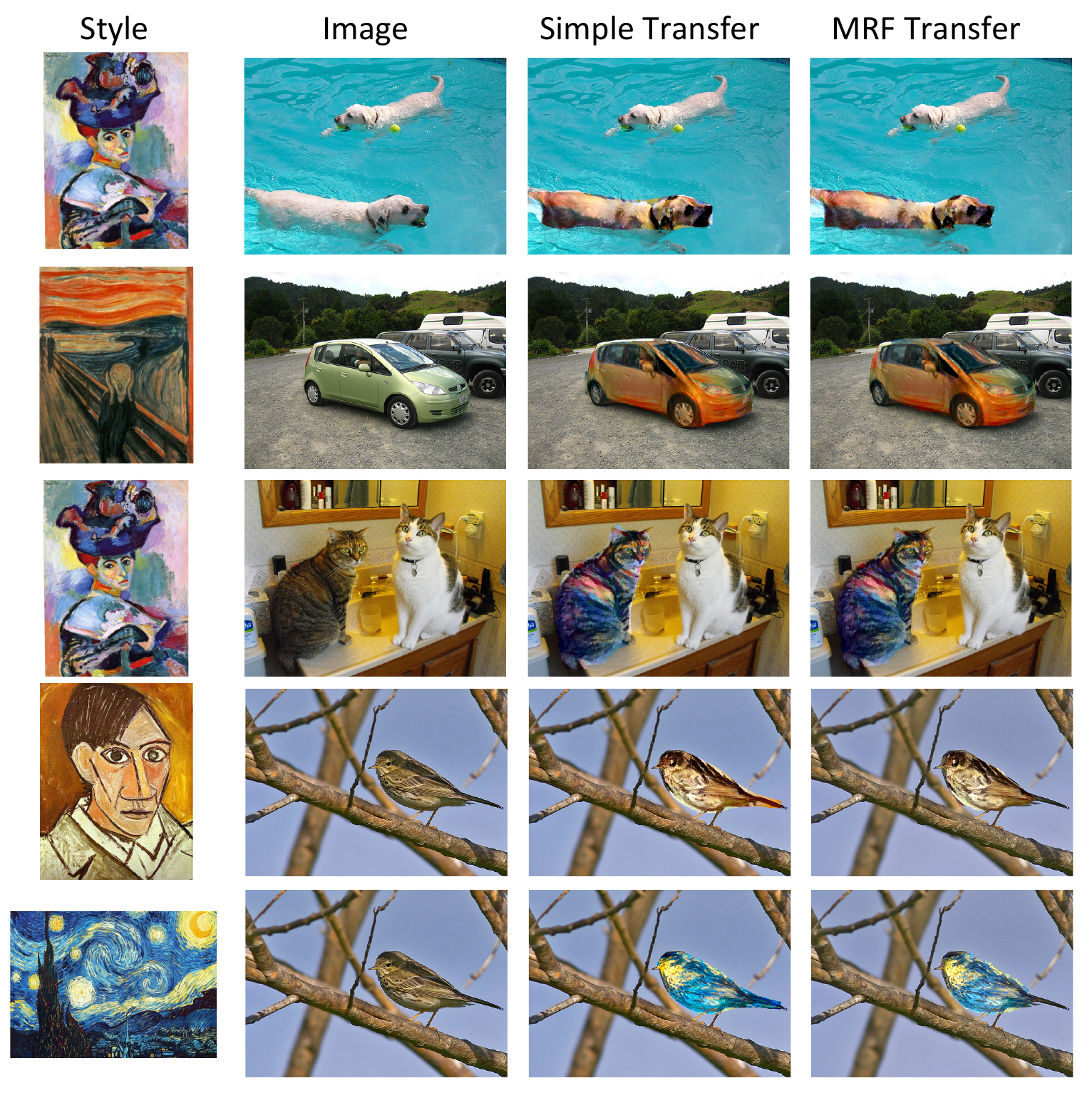}

   \caption{Results on a variety of real world images. Observe the artifacts in the non-MRF transfer 
   images at the right ear of the cat, the roof of the car, and the mouth of the dog. The MRF is particularly effective at smoothing the tail of the bird.}
   \label{images}
 \end{figure*}

Since there is no prior work on style transfer using segmentation masks, we show results for a simple mask transfer scheme and for MRF based blending. As mentioned before, the mask is generated by our deep instance aware segmentation algorithm. In Fig. \ref{images} the second column shows the input image. A simple mask transfer scheme where we overlay the style computed on the whole image onto the mask is presented in the third column. In the last column, we present results after jointly blending the input image with the stylized image. It can be clearly seen in the car image (at pixels on the top), that using an MRF based approach for blending helps to improve object level semantic style transfer. It improves results when the colors are not consistent between the image and stylized image at the semantic boundaries.  This is especially noticeable near the bird's tail.

\section{Conclusion}
We presented a method for transferring artistic style onto a single object within an image.  The proposed method combines style transfer with semantic segmentation, and blends results together using an MRF.   We presented results of the method, and demonstrate the improvement in object boundaries afforded by the MRF model. Future work will focus on ways to integrate the stages of the algorithm so that a single deep network can perform them in one shot.

\nocite{goldstein2010high}
\nocite{goldstein2015adaptive}


\bibliographystyle{IEEEbib}
\bibliography{starterBib.bib}

\begin{thebibliography}{10}

\bibitem{gatys2016image}
Leon~A Gatys, Alexander~S Ecker, and Matthias Bethge,
\newblock ``Image style transfer using convolutional neural networks,''
\newblock in {\em Proceedings of the IEEE Conference on Computer Vision and
  Pattern Recognition}, 2016, pp. 2414--2423.

\bibitem{gatys2015neural}
Leon~A Gatys, Alexander~S Ecker, and Matthias Bethge,
\newblock ``A neural algorithm of artistic style,''
\newblock {\em arXiv preprint arXiv:1508.06576}, 2015.

\bibitem{gatys2015texture}
Leon Gatys, Alexander~S Ecker, and Matthias Bethge,
\newblock ``Texture synthesis using convolutional neural networks,''
\newblock in {\em Advances in Neural Information Processing Systems}, 2015, pp.
  262--270.

\bibitem{gatys2016preserving}
Leon~A Gatys, Matthias Bethge, Aaron Hertzmann, and Eli Shechtman,
\newblock ``Preserving color in neural artistic style transfer,''
\newblock {\em arXiv preprint arXiv:1606.05897}, 2016.

\bibitem{johnson2016perceptual}
Justin Johnson, Alexandre Alahi, and Li~Fei-Fei,
\newblock ``Perceptual losses for real-time style transfer and
  super-resolution,''
\newblock {\em arXiv preprint arXiv:1603.08155}, 2016.

\bibitem{li2016combining}
Chuan Li and Michael Wand,
\newblock ``Combining markov random fields and convolutional neural networks
  for image synthesis,''
\newblock {\em arXiv preprint arXiv:1601.04589}, 2016.

\bibitem{girshick2015fast}
Ross Girshick,
\newblock ``Fast r-cnn,''
\newblock in {\em Proceedings of the IEEE International Conference on Computer
  Vision}, 2015, pp. 1440--1448.

\bibitem{ren2015faster}
Shaoqing Ren, Kaiming He, Ross Girshick, and Jian Sun,
\newblock ``Faster r-cnn: Towards real-time object detection with region
  proposal networks,''
\newblock in {\em Advances in neural information processing systems}, 2015, pp.
  91--99.

\bibitem{he2014spatial}
Kaiming He, Xiangyu Zhang, Shaoqing Ren, and Jian Sun,
\newblock ``Spatial pyramid pooling in deep convolutional networks for visual
  recognition,''
\newblock in {\em European Conference on Computer Vision}. Springer, 2014, pp.
  346--361.

\bibitem{pinheiro2015learning}
Pedro~O Pinheiro, Ronan Collobert, and Piotr Dollar,
\newblock ``Learning to segment object candidates,''
\newblock in {\em Advances in Neural Information Processing Systems}, 2015, pp.
  1990--1998.

\bibitem{dai2015instance}
Jifeng Dai, Kaiming He, and Jian Sun,
\newblock ``Instance-aware semantic segmentation via multi-task network
  cascades,''
\newblock {\em arXiv preprint arXiv:1512.04412}, 2015.

\bibitem{hariharan2014simultaneous}
Bharath Hariharan, Pablo Arbel{\'a}ez, Ross Girshick, and Jitendra Malik,
\newblock ``Simultaneous detection and segmentation,''
\newblock in {\em European Conference on Computer Vision}. Springer, 2014, pp.
  297--312.

\bibitem{boykov2001fast}
Yuri Boykov, Olga Veksler, and Ramin Zabih,
\newblock ``Fast approximate energy minimization via graph cuts,''
\newblock {\em IEEE Transactions on pattern analysis and machine intelligence},
  vol. 23, no. 11, pp. 1222--1239, 2001.

\bibitem{goldstein2010high}
Tom Goldstein and Simon Setzer,
\newblock ``High-order methods for basis pursuit,''
\newblock {\em UCLA CAM Report}, pp. 10--41, 2010.

\bibitem{goldstein2015adaptive}
Tom Goldstein, Min Li, and Xiaoming Yuan,
\newblock ``Adaptive primal-dual splitting methods for statistical learning and
  image processing,''
\newblock in {\em Advances in Neural Information Processing Systems}, 2015, pp.
  2089--2097.

\end{thebibliography}

\end{document}